# Anytime Heuristic Search

**Eric A. Hansen**                                            HANSEN@CSE.MSSTATE.EDU
*Department of Computer Science and Engineering*
*Mississippi State University*
*Mississippi State, MS 39762 USA*

**Rong Zhou**                                                    RZHOU@PARC.COM
*Palo Alto Research Center*
*3333 Coyote Hill Road*
*Palo Alto, CA 94304 USA*

## Abstract

We describe how to convert the heuristic search algorithm A* into an anytime algorithm that finds a sequence of improved solutions and eventually converges to an optimal solution. The approach we adopt uses weighted heuristic search to find an approximate solution quickly, and then continues the weighted search to find improved solutions as well as to improve a bound on the suboptimality of the current solution. When the time available to solve a search problem is limited or uncertain, this creates an anytime heuristic search algorithm that allows a flexible tradeoff between search time and solution quality. We analyze the properties of the resulting Anytime A* algorithm, and consider its performance in three domains; sliding-tile puzzles, STRIPS planning, and multiple sequence alignment. To illustrate the generality of this approach, we also describe how to transform the memory-efficient search algorithm Recursive Best-First Search (RBFS) into an anytime algorithm.

## 1. Introduction

A widely-used framework for problem-solving in artificial intelligence is heuristic search for a minimum-cost solution path in a graph. For large and complex problems, finding an optimal solution path can take a long time and a suboptimal solution that can be found quickly may be more useful. Various techniques for modifying a heuristic search algorithm such as A* to allow a tradeoff between solution quality and search time have been studied. One approach is to weight an admissible evaluation function to make it non-admissible (Pohl, 1970a, 1970b; Pearl, 1984). In the substantial literature on *weighted heuristic search*, the assumption is that the search stops as soon as the first solution is found. Analysis has focused on characterizing the tradeoff between the time it takes to find the first solution and its quality. For example, it has been shown that the cost of the first solution found will not exceed the optimal cost by a factor greater than $1+\epsilon$, where $\epsilon$ depends on the weight (Pearl, 1984; Davis, Bramanti-Gregor, & Wang, 1988). There have also been empirical studies of the tradeoff between search time and the quality of the first solution found (Gasching, 1979; Korf, 1993).

The starting point for this paper is the simple observation that there is no reason to stop a non-admissible search after the first solution is found. By continuing the search, a sequence of improved solutions can be found that eventually converges to an optimal solution. The possibility of continuing a non-admissible A* search after the first solution is found was suggested by Harris (1974), although he did not consider weighted heuristic search but a somewhat related approach called bandwidth heuristic search. We are not aware of any other mention of this idea before we proposed it as a strategy for creating an Anytime A* algorithm (Hansen, Zilberstein, & Danilchenko, 1997; Zhou & Hansen, 2002). In this paper, we discuss this strategy at length and evaluate it analytically and empirically.

We refer to this strategy as *anytime heuristic search*. Anytime algorithms are useful for problem-solving under varying or uncertain time constraints because they have a solution ready whenever they are stopped (with the possible exception of an initial time period before the first solution is found) and the quality of the solution improves with additional computation time (Zilberstein, 1996). Because heuristic search has many applications, a general method for transforming a heuristic search algorithm such as A* into an anytime algorithm could prove useful in many domains where good anytime algorithms are not otherwise available.

The paper is organized as follows. Section 2 presents our approach for transforming a weighted heuristic search algorithm into an anytime algorithm, and shows how to transform Weighted A* into an Anytime A* algorithm. To illustrate the generality of this approach, Section 3 considers Recursive Best-First Search (RBFS), which is a memory-efficient version of A*, and shows how to transform Weighted RBFS into an Anytime RBFS algorithm. Section 4 discusses related work, including a related approach to creating an Anytime A* algorithm that has been recently proposed.

## 2. Anytime A*

We begin this section with a brief review of the standard A* and Weighted A* algorithms. Then we describe how to transform Weighted A* into an Anytime A* algorithm. We analyze the theoretical properties of the resulting algorithm and evaluate its performance in three test domains; sliding-tile puzzles, STRIPS planning, and multiple sequence alignment.

### 2.1 A*

The A* algorithm (Hart, Nilsson, & Raphael, 1968) uses two lists, an Open list and a Closed list, to manage a systematic search for a minimum-cost path from a start node to a goal node in a graph. Initially, the Open list contains the start node and the Closed list is empty. At each cycle of the algorithm, the most promising open node is expanded, moved to the Closed list, and its successor nodes are inserted into the Open list. Thus, the Closed list contains those nodes that have been expanded, by generating their successor nodes, and the Open list contains those nodes that have been generated, but not yet expanded. The search terminates when a goal node is selected for expansion. A solution path can be extracted by tracing node pointers backwards from the goal node to the start node.

The order in which nodes are expanded is determined by the node evaluation function $f(n) = g(n) + h(n)$, where $g(n)$ is the cost of the best path currently known from the start node to node $n$, and $h(n)$ is a heuristic estimate of $h^*(n)$, the cost of the best path from $n$ to a goal node. The behavior of A* depends in large part on the heuristic $h(n)$ that guides the search. If $h(n)$ is *admissible*, that is, if it never overestimates $h^*(n)$, and if nodes are expanded in order of $f(n)$, then the first goal node selected for expansion is guaranteed to be optimal. A heuristic is said to be *consistent* if $h(n) \leq c(n, n') + h(n')$ for all $n$ and $n'$, where $c(n, n')$ is the cost of an edge from node $n$ to node $n'$. If $h(n)$ is consistent and nodes are expanded in order of $f(n)$, the $g$-cost of a node is guaranteed to be optimal when the node is selected for expansion, and a node is never expanded more than once. Note that consistency implies admissibility, and non-admissibility implies inconsistency.

If $h(n)$ is not consistent, it is possible for A* to find a better path to a node after the node is expanded. In this case, the improved $g$-cost of a node needs to be propagated to its descendants. The way that A* usually does this is by reopening nodes, that is, by moving a node from the Closed list to the Open list when its $g$-cost is improved. When the node is eventually reexpanded, the improved $g$-cost is propagated to its successor nodes, which may need to be reopened also. As a result, the same node can be expanded multiple times. Although rarely used in practice, various techniques have been introduced to bound the worst-case number of node reexpansions (Bagchi & Mahanti, 1983; Bagchi & Srimani, 1985).

## 2.2 Weighted A*

For difficult search problems, A* may take too long to find an optimal solution, and an approximate solution that is found relatively quickly can be more useful. Beginning with Pohl (1970a, 1970b), many researchers have explored the effect of weighting the terms $g(n)$ and $h(n)$ in the node evaluation function differently, in order to allow A* to find a bounded-optimal solution with less computational effort. In this approach, called *Weighted A* * (WA*), the node evaluation function is defined as $f'(n) = g(n) + w \times h(n)$, where the weight $w \geq 1.0$ is a parameter set by the user. Sometimes the node evaluation function is defined as $f'(n) = (1-w') \times g(n) + w' \times h(n)$, but this is equivalent to $f'(n) = g(n) + w \times h(n)$ when $w' = \frac{w}{1+w}$. We use the notation $f'(n)$ to distinguish a weighted evaluation function from the unweighted $f(n)$. If $w > 1.0$, the search is not admissible and the (first) solution found may not be optimal, although it is usually found much faster. If $h(n)$ is admissible, the suboptimality of the solution found by weighted heuristic search is bounded: the solution cost cannot be greater than the optimal cost by more than a factor of $w$ (Davis et al., 1988). Such a solution is said to be $\epsilon$-*admissible* where $\epsilon = w - 1.0$. A weighted heuristic accelerates search for a solution because it makes nodes closer to a goal seem more attractive, giving the search a more depth-first aspect and implicitly adjusting a tradeoff between search effort and solution quality. Weighted heuristic search is most effective for search problems with close-to-optimal solutions, and can often find a close-to-optimal solution in a small fraction of the time it takes to find an optimal solution.

Some variations of weighted heuristic search have been studied. An approach called *dynamic weighting* adjusts the weight with the depth of the search (Pohl, 1973; Koll & Kaindl, 1992). Another approach uses a weighted heuristic to identify a subset of open nodes that can be expanded without loss of $\epsilon$-admissibility; from this subset, it selects the node to expand next based on other criteria (Pearl & Kim, 1982; Davis et al., 1988). Weighted heuristic search has been used with other search algorithms besides A*, including memory-efficient versions of A* such as IDA* and RBFS (Korf, 1993), as well as Learning Real-Time A* (LRTA*) (Shimbo & Ishida, 2003), and heuristic search algorithms for AND/OR graphs (Chakrabarti, Ghosh, & DeSarkar, 1988; Hansen & Zilberstein, 2001).

## 2.3 Anytime Weighted A*

We now consider how Weighted A* can be transformed into an anytime algorithm that finds a sequence of improved solutions and eventually converges to an optimal solution. The transformation is an example of a more general approach for transforming a search algorithm that explores nodes in best-first order, such as A*, into an anytime algorithm. This approach consists of the following three changes.

1. A non-admissible evaluation function, $f'(n) = g(n) + h'(n)$, where the heuristic $h'(n)$ is not admissible, is used to select nodes for expansion in an order that allows good, but possibly suboptimal, solutions to be found quickly.

2. The search continues after a solution is found, in order to find improved solutions.

3. An admissible evaluation function (i.e., a lower-bound function), $f(n) = g(n) + h(n)$, where $h(n)$ is admissible, is used together with an upper bound on the optimal solution cost (given by the cost of the best solution found so far), in order to prune the search space and detect convergence to an optimal solution.

In this paper, we use a weighted heuristic to create the non-admissible evaluation function that guides the search. That is, we assume that we have an admissible heuristic $h(n)$, and use it to create a weighted heuristic $h'(n) = w \times h(n)$. But this three-step approach for creating an anytime heuristic search algorithm can use any non-admissible heuristic that helps A* find an approximate solution quickly; it is not limited to weighted heuristic search. When the general approach is used to transform A* into an anytime algorithm, we call the resulting algorithm *Anytime A\**. In the special case in which the non-admissible evaluation function is a weighted heuristic, we call the algorithm *Anytime Weighted A\** or *Anytime WA\**.

Algorithm 1 shows high-level pseudocode for Anytime WA*. (Some details that are unaffected by the transformation of WA* into an anytime algorithm, such as extracting the solution path, are omitted.) Note that our implementation of Anytime A* tests whether a node is a goal node as soon as the node is generated and not when it is selected for expansion, as in A*, since this can improve the currently available solution more quickly.

Besides continuing the search after the first solution is found, Anytime WA* uses bounds to prune the search space. The sequence of improved solutions found by Anytime WA* provides a sequence of improved upper bounds on the optimal solution cost. Anytime WA* tests whether the $f$-cost of each newly-generated node is less than the current upper bound. If not, the node is not inserted in the Open list since it cannot lead to an improved solution. By not inserting suboptimal nodes into the Open list, the memory requirements of the algorithm are reduced.[1] Each time an improved solution is found and the upper bound decreases, it is possible that some nodes already in the Open list may have an $f$-cost equal to or greater than the new upper bound. Although these nodes could be immediately

---

1. The possibility of using bounds on the optimal solution cost to reduce the number of nodes stored in the Open list has been suggested at least twice before in the literature. Harris (1974, p. 219) points out that this can be done when a *bandwidth heuristic* is used to guide the search, which is a heuristic with error bounded by an additive constant. Such heuristics may not be easy to obtain, however. Ikeda and Imai (1994) describe an *Enhanced A\** algorithm that uses a previously-computed upper bound to limit the number of nodes stored in the Open list. We compare Enhanced A* to Anytime WA* in Section 2.4.3.

---

**Algorithm 1**: Anytime-WA*

---

**Input**: A start node *start*
**Output**: Best solution found and error bound
**begin**
    $g(start) \leftarrow 0,\ f(start) \leftarrow h(start),\ f'(start) \leftarrow w \times h(start)$
    $OPEN \leftarrow \{start\},\ CLOSED \leftarrow \emptyset,\ incumbent \leftarrow \textbf{nil}$
    **while** $OPEN \neq \emptyset$ **and not** *interrupted* **do**
        $n \leftarrow \arg\min_{x \in OPEN} f'(x)$
        $OPEN \leftarrow OPEN \setminus \{n\}$
        **if** $incumbent = \textbf{nil}$ **or** $f(n) < f(incumbent)$ **then**
            $CLOSED \leftarrow CLOSED \cup \{n\}$
            **foreach** $n_i \in Successors(n)$ **such that** $g(n) + c(n, n_i) + h(n_i) < f(incumbent)$ **do**
                **if** $n_i$ is a goal node **then**
                    $f(n_i) \leftarrow g(n_i) \leftarrow g(n) + c(n, n_i),\ incumbent \leftarrow n_i$
                **else if** $n_i \in OPEN \cup CLOSED$ **and** $g(n_i) > g(n) + c(n, n_i)$ **then**
                    $g(n_i) \leftarrow g(n) + c(n, n_i),\ f(n_i) \leftarrow g(n_i) + h(n_i),\ f'(n_i) \leftarrow g(n_i) + w \times h(n_i)$
                    **if** $n_i \in CLOSED$ **then**
                        $OPEN \leftarrow OPEN \cup \{n_i\}$
                        $CLOSED \leftarrow CLOSED \setminus \{n_i\}$
                **else**
                    $g(n_i) \leftarrow g(n) + c(n, n_i),\ f(n_i) \leftarrow g(n_i) + h(n_i),\ f'(n_i) \leftarrow g(n_i) + w \times h(n_i)$
                    $OPEN \leftarrow OPEN \cup \{n_i\}$
    **if** $OPEN = \emptyset$ **then** $error \leftarrow 0$
    **else** $error \leftarrow f(incumbent) - \min_{x \in OPEN} f(x)$
    **output** incumbent solution and error bound
**end**

---

removed from the Open list, this incurs overhead for searching the Open list every time the upper bound decreases, and so this step is not included in the pseudocode. (Of course, if Anytime WA* is close to running out of memory, the overhead for searching through the Open list for sub-optimal nodes that can be deleted might be justified by the need to recover memory.) The algorithm shown in the pseudocode simply tests the $f$-cost of each node before expanding it. If the $f$-cost is equal to or greater than the current upper bound, the node is not expanded. This implies a related test for convergence to an optimal solution: if the Open list is empty, the currently available solution must be optimal.

Anytime WA* requires more node expansions than A* to converge to an optimal solution, for two reasons. First, use of a weighted heuristic allows it to expand more distinct nodes than A*. Second, a weighted heuristic is inconsistent and this means it is possible for a node to have a higher-than-optimal $g$-cost when it is expanded. If a better path to the same node is later found, the node is reinserted in the Open list so that the improved $g$-cost can be propagated to its descendants when the node is reexpanded. This means that Anytime WA* can expand the same node multiple times.

Before considering the empirical performance of Anytime WA*, we discuss two important theoretical properties of the algorithm: convergence to an optimal solution, and a bound on the suboptimality of the currently available solution.

### 2.3.1 CONVERGENCE

An admissible evaluation function, $f(n)$, gives a lower bound on the cost of any solution path that is an extension of the current best path to node $n$. Let *incumbent* denote the goal node corresponding to the best solution found so far. Then $f(incumbent)$ is an upper bound on the cost of an optimal solution. Clearly, there is no reason to expand a node that has an $f$-cost (i.e., a lower bound) greater than or equal to the current upper bound, $f(incumbent)$, since it cannot lead to an improved solution. Thus we have the following convergence test for Anytime WA*, and, more generally, for any anytime version of A*: the best solution found so far is optimal if there are no unexpanded nodes on the search frontier (i.e., in the Open list) with an $f$-cost less than $f(incumbent)$.

We prove the following theorem under the standard assumptions that a search graph has a finite branching factor and a minimal edge cost $c > 0$. We also assume that a solution exists and that $h(n) \geq 0$ for all nodes $n$.

**Theorem 1** *Anytime WA\* always terminates and the last solution it finds is optimal.*

**Proof**: First we show that the algorithm cannot terminate before an optimal solution is found. Suppose that the algorithm terminates before finding an optimal solution which has cost $f^*$. The sequence of upper bounds used during execution of the algorithm is $b_0, b_1, ...., b_k$, where $b_0 = \infty$ (the upper bound before any solution is found), $b_1$ is the cost of the first solution found, and $b_k$ is the cost of the last solution found. We know that,

$$b_0 > b_1 > ... > b_k > f^*,$$

where the last inequality holds under the assumption that the algorithm terminates before finding an optimal solution, that is, with a suboptimal solution.

Now consider an optimal path leading from the initial state to a goal state. Under the assumption that this optimal solution path was not found, there must be some node $n$ along this path that was generated but not expanded. That is only possible if $g(n) + h(n) \geq b_k$. But by the admissibility of $h$, we know that

$$g(n) + h(n) \leq f^*,$$

and therefore

$$\forall i \; : \; g(n) + h(n) \leq f^* < b_i.$$

From this contradiction, it follows that the algorithm cannot terminate before an optimal solution is found.

Next we show that the algorithm always terminates. We have already proved that before an optimal solution is found, the Open list must include some node $n$ for which $g(n) + h(n) \leq f^*$. Hence,

$$f'(n) = g(n) + w \times h(n) \leq w \times (g(n) + h(n)) \leq w \times f^*.$$

This establishes an upper bound on the $f$-cost of any node that can be expanded by Anytime WA* before an optimal solution is found. Because there are a finite number of nodes for which $f(n) \leq w \times f^*$, the algorithm must run a bounded number of steps before an optimal solution is found. Once an optimal solution is found, the algorithm will not expand any node with an $f$-cost that is greater than or equal to $f^*$. Because the number of nodes for which $f(n) \leq f^*$ is also finite, the algorithm must eventually terminate. $\square$

### 2.3.2 Error bound

An important property of this approach to creating an Anytime A* algorithm is that it refines both an upper and lower bound on the optimal cost of a solution. The upper bound is the $f$-cost of the best solution found so far, and is decreased when an improved solution is found. The lower bound is the least $f$-cost of any currently open node, and is increased when all nodes with the smallest $f$-cost are expanded.

Although it is obvious that the cost of the best solution found so far is an upper bound, the claim that the least $f$-cost of any currently open node is a lower bound on the optimal solution cost requires some justification. First note that if the currently available solution is not optimal, an optimal solution path must pass through some currently open node. Although the $f$-cost of an open node is not necessarily a lower bound on the best solution path that passes through that node, since the $g$-cost of an open node may be suboptimal, it is a lower bound on the cost of any solution path that is an extension of the current path to that node. Since any improved path to an open node (resulting in an improved $g$-cost) must be an extension of some already-found path to another currently open node with lower $f$-cost, the least $f$-cost of any currently open node must be a lower bound on the cost of an optimal solution path. In other words, the least $f$-cost of any currently open node is a lower bound on the optimal solution cost for the same reason that it is a lower bound on the optimal solution cost in branch-and-bound tree search.

These upper and lower bounds approach each other as the search progresses until they meet upon convergence to an optimal solution. (Figure 3 shows an example of how the bounds approach each other.) Before an optimal solution is found, a bound on the difference between the cost of the currently available solution and the optimal solution cost is given by the difference between the upper and lower bounds. This error bound can be expressed as an approximation ratio, such that $\frac{f(incumbent)}{f^*} \leq \frac{f(incumbent)}{f^L}$, where $f^L$ denotes the lower bound on the optimal solution cost, $f(incumbent)$ denotes the upper bound, and $f^*$ denotes the optimal solution cost. Thus, Anytime A* can be viewed as an anytime algorithm in two respects. It improves a solution over time, and it also improves a bound on the suboptimality of the currently available solution.

## 2.4 Performance and evaluation

We next consider the empirical performance of Anytime WA* in solving a range of search problems. Its effectiveness depends on the weight used, and how the weight affects search performance depends on characteristics of both the problem and the heuristic. We set the weight based on a combination of knowledge of the search problem and trial and error. We use the same weight from start to end of the search, which has the advantage of simplicity. It also shows that the technique in its simplest form leads to good performance. It is possible to change the weight during the search, and weight adjustment together with other methods of search control has the potential to improve performance further. We postpone discussion of this to Section 4.1 where we discuss a variant of Anytime A* that has recently been proposed.

For search problems with unit edge costs, many nodes have the same $f$-cost and the tie-breaking rule used by a systematic search algorithm such as A* has a significant effect on the number of nodes actually expanded. It is well-known that A* achieves best performance

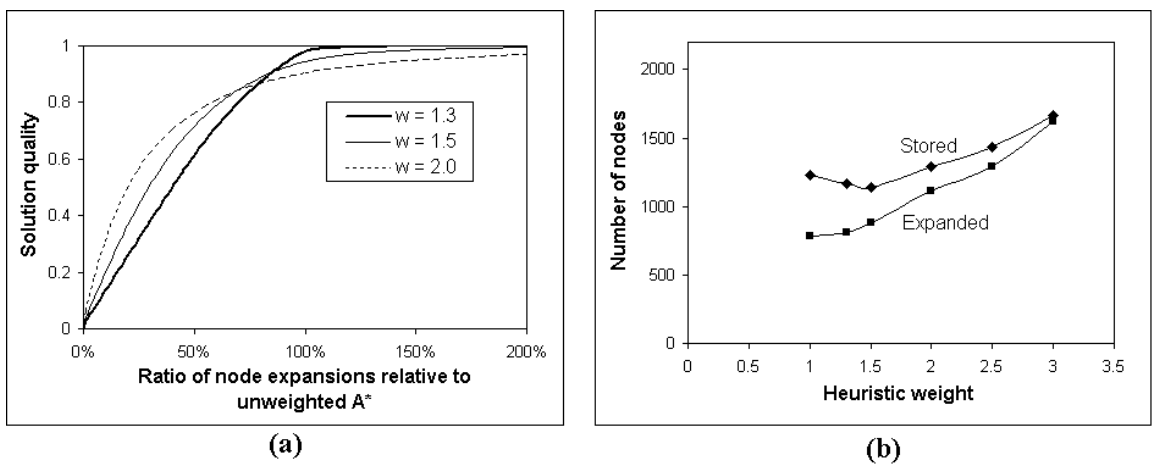

**(a)**          **(b)**

Figure 1: (a) Performance profiles for Anytime WA* using three different weights, averaged over all instances of the Eight Puzzle. (b) Average number of nodes stored and expanded by Anytime WA* over all instances of the Eight Puzzle.

when it breaks ties in favor of nodes with the least $h$-cost. In all of the experimental comparisons reported in this paper, A* uses this tie-breaking rule. We note that Anytime WA* can achieve similar tie-breaking behavior without applying the same rule because using even a very small weight has the effect of breaking ties in favor of nodes with the least $h$-cost. Moreover, Anytime WA* usually finds an optimal solution before it can prove that it is optimal (that is, before it expands all nodes with an $f$-cost less than the optimal $f$-cost). As a result, it usually does not expand non-goal nodes with an $f$-cost equal to the optimal solution cost. For consistency in experimental comparison, our implementation of Anytime WA* uses the same rule that A* uses of breaking ties in favor of nodes with the least $h$-cost. In practice, this tie-breaking rule can be omitted when implementing Anytime WA* in order to reduce run-time overhead.

### 2.4.1 Sliding-tile puzzle

The first test domain we consider is a traditional benchmark that lets us illustrate this technique on a simple and well-understood example. Figure 1(a) shows performance profiles for Anytime WA* using three different weights, averaged over all instances of the Eight Puzzle.[2] (*Performance profiles* are commonly used to model the performance of anytime algorithms, and show how expected solution quality improves as a function of computation time. For these problems, we define the quality of a solution as $1 - \frac{f - f^*}{f^*}$.) A weight of 1.3 seems to result in the best overall performance among these three weights, although it does not dominate the other performance profiles for all running times.

---

2. By all instances, we mean all possible starting states and a fixed goal state. The goal state has the blank in the upper left corner and the tiles arranged in numerical order, left-to-right and then top-down. We use the Manhattan distance heuristic.

| Instance | Len | A* | | | AWA* (weight = 2) | | | |
|---|---|---|---|---|---|---|---|---|
| | | Stored | Exp | Secs | Stored | Exp | Opt % | Secs |
| Blocks-8 | 14 | 426,130 | 40,638 | 5.2 | 41,166 | 41,099 | 0.2% | 3.8 |
| Logistics-6 | 25 | 364,846 | 254,125 | 4.0 | 254,412 | 254,748 | 6.2% | 3.7 |
| Satellite-6 | 20 | 3,270,195 | 2,423,373 | 151.5 | 2,423,547 | 2,423,566 | 14.3% | 138.8 |
| Freecell-3 | 18 | 5,992,688 | 2,693,235 | 170.0 | 2,695,321 | 2,705,421 | 1.7% | 146.2 |
| Psr-46 | 34 | 7,464,850 | 7,141,461 | 343.2 | 7,148,048 | 7,175,275 | 69.0% | 348.0 |
| Depots-7 | 21 | 21,027,257 | 7,761,661 | 367.8 | 7,773,830 | 7,772,091 | 0.5% | 249.1 |
| Driverlog-11 | 19 | 22,344,515 | 6,693,096 | 407.0 | 6,702,570 | 6,699,143 | 1.4% | 281.6 |
| Elevator-12 | 40 | 12,748,119 | 12,734,334 | 560.6 | 12,734,636 | 12,829,775 | 98.6% | 569.7 |

Table 1: Comparison of A* and AWA* on eight benchmark problems from the biennial Planning Competitions.

Figure 1(b) shows how many nodes Anytime WA* stores and expands before it converges to an optimal solution, using different weights. (Again, by "converges to an optimal solution", we mean that the lower and upper bounds meet and the algorithm has proved that the solution is optimal.) Using a weight of 1.3, the average increase in the number of nodes expanded by Anytime WA* is very slight compared to the number of nodes expanded by unweighted A*. Figure 1(b) also shows that Anytime WA* using a weight of 1.3 or 1.5 stores fewer nodes than unweighted A*. For these weights, the reduction in memory requirements due to using an upper bound to prune the Open list is greater than the increase in memory requirements due to expanding more distinct nodes.

### 2.4.2 STRIPS planning

In recent years, there has been considerable interest in using heuristic search for domain-independent STRIPS planning. Influential examples of this approach are the HSP and HSPr planners of Bonet and Geffner (2001), which have performed well in the biennial planning competitions sponsored by the International Conference on Automated Planning and Scheduling (Long & Fox, 2003). HSP solves STRIPS planning problems using A* to search forward from the start state to the goal, and HSPr uses A* to search backwards from the goal to the start state, which has the advantage that it allows the heuristic to be computed more efficiently. Because many of the benchmark planning problems used in the planning competition are difficult to solve optimally, WA* is often used to find suboptimal solutions in a reasonable amount of time.

Using Bonet's publicly-available implementation of HSPr, we compared the performance of A* and Anytime WA* on benchmark problems from previous planning competitions that could be solved optimally by A*, using the domain-independent and admissible *max-pair* heuristic described by Haslum and Geffner (2000). We used a weight of 2.0 in our experiments. For all instances, Anytime WA* converged to an optimal solution using less memory than A*. For most (but not all) instances, it also took less time. Table 1 compares the performance of A* and Anytime WA* (AWA*) on the hardest solvable instances of eight of these planning problems.[3] The CPU time is relatively long for the number of nodes generated and stored due to significant overhead for generating a node and computing its heuristic in a domain-independent way. The Blocks and Driverlog domains have the largest branching factors, and thus the space savings from using an upper bound to prevent insertion of suboptimal nodes in the Open list are greatest in these domains. In no domain did Anytime WA* expand as many as 1% more nodes than A*, and usually the increased percentage of node expansions is a fraction of this.

| | AWA* (weight = 5) | | | AWA* (weight = 10) | | |
|---|---|---|---|---|---|---|
| Instance | Exp | Opt % | Secs | Exp | Opt % | Secs |
| Blocks-8 | 42,293 | 0.2% | 3.9 | 42,293 | 0.2% | 3.9 |
| Logistics-6 | 274,047 | 4.6% | 3.8 | 312,726 | 11.0% | 4.3 |
| Satellite-6 | 2,458,452 | 8.9% | 138.7 | 2,585,074 | 13.2% | 144.8 |
| Freecell-3 | > 35,639,419 | N/A | > 2,207.6 | > 73,712,127 | N/A | > 4,550.5 |
| Psr-46 | 7,310,349 | 10.0% | 350.6 | 7,623,007 | 4.8% | 365.1 |
| Depots-7 | 7,902,183 | 0.4% | 250.7 | 8,115,603 | 1.9% | 254.7 |
| Driverlog-11 | 6,814,696 | 1.2% | 281.1 | 7,674,956 | 18.0% | 322.4 |
| Elevator-12 | 12,851,075 | 76.0% | 557.5 | 13,145,547 | 21.4% | 568.0 |

Table 2: Performance of AWA* with weights 5 and 10 on eight benchmark problems from the biennial Planning Competitions.

The column labeled "Opt %" shows how soon Anytime WA* finds what turns out to be an optimal solution. The percentage is the number of node expansions before finding the optimal solution out of the total number of node expansions until convergence. This provides a very rough measure of the anytime performance of the algorithm. It shows that in most domains, Anytime WA* finds what turns out to be an optimal solution very quickly and spends most of its search time proving that the solution is optimal. However in two domains (Psr-46 and Elevator-12), Anytime WA* did not find any solution until relatively late. In both of these domains, solutions were found sooner when the weight was increased. Table 2 shows the performance of Anytime WA* using weights of 5 and 10. Using these higher weights, anytime performance is better for the last two problems, especially Elevator-12, although worse for some of the others. Even with weights of 5 and 10, Anytime WA* tends to outperform A* in solving the first five problems. The sixth problem, Freecell-3, is different. With a weight of 5, Anytime WA* cannot find any solution before running out of memory. With a weight of 10, the number of stored nodes is the same (since it exhausts the same amount of memory) but the number of expanded nodes (and the CPU time) more than doubles because there are more node reexpansions as the weight increases. These results clearly show that the effect of the weight on search performance can vary with the domain. Given this variability, some trial and error in selecting the weight appears inevitable. But if an appropriate weight is used, Anytime Weighted A* is consistently beneficial.

---

3. All our experiments were performed on an Intel Xeon 3.0GHz processor with 2MB of L2 cache and 2GB of RAM

### 2.4.3 Multiple sequence alignment

Alignment of multiple DNA or protein sequences plays an important role in computational molecular biology. It is well-known that this problem can be formalized as a shortest-path problem in an $n$-dimensional lattice, where $n$ is the number of sequences to be aligned (Carillo & Lipman, 1988; Yoshizumi, Miura, & Ishida, 2000). A* can outperform dynamic programming in solving this problem by using an admissible heuristic to limit the number of nodes in the lattice that need to be examined to find an optimal alignment (Ikeda & Imai, 1999). However, a challenging feature of this search problem is its large branching factor, which is equal to $2^n - 1$. When A* is applied to this problem, the large branching factor means the Open list can be much larger than the Closed list, and the memory required to store the Open list becomes a bottleneck of the algorithm.

Two solutions to this problem have been proposed in the literature. Yoshizumi et al. (2000) describe an extension of A* called *A\* with Partial Expansion* (PEA*). Instead of generating all successors of a node when it is expanded, PEA* inserts only the most promising successors into the Open list. The "partially expanded" node is then reinserted into the Open list with a revised $f$-cost equal to the least $f$-cost of its successors that have not yet been generated, so that the node can be reexpanded later. Use of this technique significantly reduces the size of the Open list, and PEA* can solve larger multiple sequence alignment problems than A*. Unfortunately, the reduced space complexity of PEA* is achieved at the cost of node reexpansion overhead. The tradeoff between space and time complexity is adjusted by setting a "cutoff value" $C$, which implicitly determines how many successor nodes to add to the Open list at a time.

Another way to reduce the size of the Open list is not to insert nodes into the Open list if their $f$-cost is equal to or greater than a previously established upper bound on the cost of an optimal solution, since such nodes will never be expanded by A*. This approach was proposed by Ikeda and Imai (1999), who call it *Enhanced A\** (EA*). They suggest that one way to obtain an upper bound is to use the solution found by Weighted A* search with a weight $w > 1$, although they did not report experimental results using this technique.

Our anytime algorithm provides a third approach to reducing the size of the Open list. We also use Weighted A* to quickly find a solution that provides an upper bound for pruning the Open list. But because the first solution found may not be optimal, the weighted search is continued in order to find a sequence of improved solutions that eventually converges to an optimal solution. This provides a sequence of improved upper bounds that can further reduce the size of the Open list.

Figure 2 compares the performance of Anytime WA* (AWA*) to the performance of A* with Partial Expansion and Enhanced A* in aligning five sequences from a set of dissimilar (and thus difficult to align) sequences used in earlier experiments (Kobayashi & Imai, 1998). The cost function is Dayhoff's PAM-250 matrix with a linear gap cost of 8. The admissible heuristic is the standard pairwise alignment heuristic, and the (almost negligible) time needed to compute the heuristic is included in the running time of the search.

All three algorithms require much less memory than standard A* in solving this problem. We found that a good weight for Anytime WA* in solving this test problem is $\frac{100}{99}$, that is, the $g$-cost is weighted by 99 and the $h$-cost is weighted by 100. (Because the cost function for multiple sequence alignment is integer-valued, we use a weighting scheme that preserves integer $f$-costs to allow more efficient integer-valued arithmetic.) To create an upper bound for Enhanced A*, we ran Weighted A* with the same weight of $\frac{100}{99}$ and used the cost of the first solution found as the upper bound.

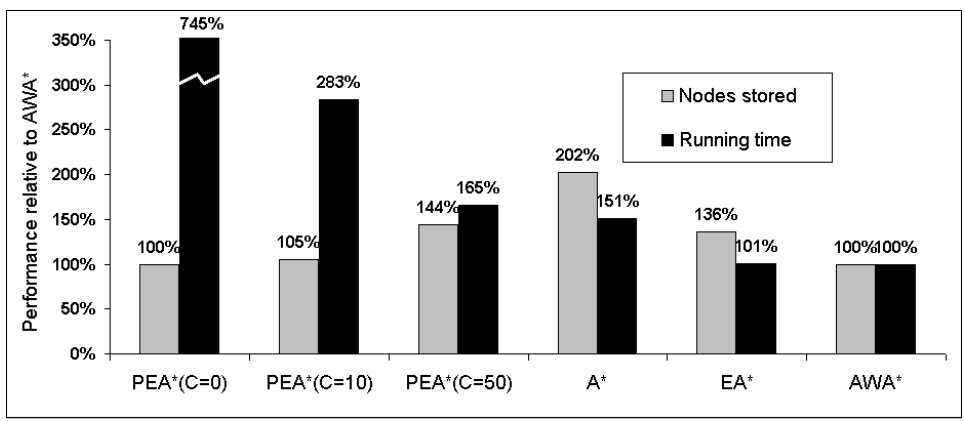

Figure 2: Average performance of search algorithms in aligning sets of 5 sequences from Kobayashi and Imai (1998).

Figure 2 shows that Anytime WA* runs more than seven times faster and stores about the same number of nodes as PEA* with a cutoff of $C = 0$. When PEA* uses a cutoff of $C = 50$, it stores 44% more nodes than Anytime WA* and still runs 65% slower on average. Although Enhanced A* runs about as fast as Anytime WA*, it stores 36% more nodes. Anytime WA* stores fewer nodes because continuation of weighted search results in discovery of improved solutions that provide tighter upper bounds for pruning the Open list. In summary, Anytime WA* not only outperforms standard A* in solving this problem, it performs better than two state-of-the-art enhancements of A* that were specifically designed for this problem.

Figure 3 illustrates the behavior of Anytime WA* by showing how the upper and lower bounds gradually converge. Notice that Anytime WA* finds an optimal solution after only 10% of the total search time, and spends the remaining 90% of the time proving that the solution is optimal, at which point it converges. Compared to Partial Expansion A* and Enhanced A*, an important advantage of Anytime WA* is that it finds a suboptimal alignment quickly and then continues to improve the alignment with additional computation time. Thus, it offers a tradeoff between solution quality and computation time that can prove useful when finding an optimal alignment is infeasible.

The weight that we found worked well for this problem may seem surprisingly small, and one might suspect that a weight this small has little or no effect on the order in which nodes with different $f$-costs are expanded, and serves primarily as a tie-breaking rule for nodes with the same $f$-cost but different $h$-costs. Because our implementations of A* and Anytime WA* both break ties in favor of nodes with the least $h$-cost, however, the weight has no effect on tie breaking in our experiments.

There are a couple of reasons why such a small weight is effective for this search problem. First, the search graph for multiple sequence alignment has non-uniform edge costs. As a result, the range of $f$-costs is much greater than for our other test problems, which have unit edge costs. Second, the $f$-costs and $h$-costs are *much* larger for this problem than for our other test problems – in part, because the edge costs are larger (given the cost function we used), and, in part, because the search space is deeper. (The protein sequences being aligned have an average length of about 150, and this means the search is at least this deep.) For this search problem, the optimal $f$-costs are around $50,000$. Because the pairwise alignment heuristic used in solving this problem is very accurate, the largest $h$-costs are also around $50,000$. Given $h$-costs this large and a wide range of $f$-costs, a weight of $\frac{100}{99}$ can have a significant effect on the order of node expansions. This serves to illustrate how the appropriate weight for Anytime WA* depends on characteristics of the search problem.

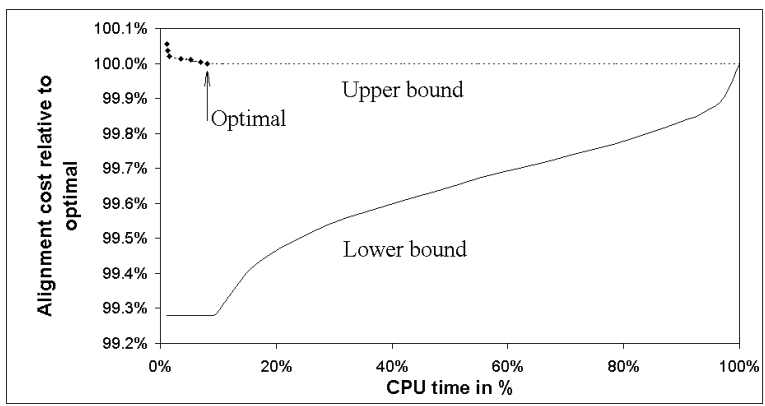

Figure 3: Convergence of bounds for Anytime WA* in aligning sets of 5 sequences from Kobayashi and Imai (1998).

### 2.4.4 DISCUSSION

Our results show that Anytime WA* is effective for a wide range of search problems. In general, it is effective for a search problem whenever Weighted A* is effective. As others have observed, Weighted A* can usually find a solution much faster than A* because A* spends most of its time discriminating between close-to optimal solutions in order to determine which is optimal (Pearl, 1984, p. 86). Indeed, our test results show that Anytime WA* often finds what turns out to be an optimal solution relatively quickly, and spends most of its search time proving that the solution is optimal.

One of the surprising results of our experiments is that Anytime WA* using an appropriate weight can sometimes converge to an optimal solution using less memory and even less time than A*. This is surprising because it is well-known that A* using a consistent heuristic is "optimally efficient" in terms of the number of nodes expanded (Dechter & Pearl, 1985). However it is not necessarily optimally efficient by other measures of search complexity, including memory requirements and running time. Anytime WA* is sometimes more efficient by these other measures of search performance, even though it requires more node expansions to find a provably optimal solution. The reason for this is that the improved solutions found by the anytime approach provide upper bounds that can be used to reduce the number of nodes stored in the Open list. The resulting savings, both in memory and in time overhead for managing the Open list, are sometimes greater than the additional overhead of increased node expansions.

In our experiments, we used relatively low weights that result in fast convergence as well as good anytime performance. This shows that A* can be transformed into an anytime algorithm in exchange for little or no delay in convergence to an optimal solution. This does not mean we recommend that the weight used by Anytime WA* should *always* be set low enough to minimize memory use or the time it takes to find a provably optimal solution. For some search problems, it could be an advantage to use higher weights in an attempt to find approximate solutions more quickly. In most cases, increasing the weight used by Anytime WA* allows an approximate solution to be found sooner, but increases the number of node expansions before convergence to an optimal solution. In the end, the "best" weight depends on preferences about time-quality tradeoffs.

We have focused on how to use weighted heuristic search to create an anytime heuristic search algorithm. But in fact, any non-admissible heuristic could be used to guide an Anytime A* algorithm, as pointed out at the beginning of Section 2.3. It is possible (and even seems likely) that a more informative, but inadmissible, heuristic could sometimes lead to better anytime search performance than a weighted admissible heuristic. In this case, Anytime A* would use two heuristics – a non-admissible heuristic to select the order of node expansions, and another, admissible heuristic, to prune the search space and detect convergence to an optimal solution. This is an interesting direction for further exploration. Our contribution in this paper is to show that an approach as simple as weighting an admissible heuristic creates a very effective anytime algorithm for many search problems.

## 3. Anytime RBFS

It is well-known that the scalability of A* is limited by the memory required to store the Open and Closed lists. This also limits the scalability of Anytime A*. Several variants of A* have been developed that use less memory, including algorithms that require only linear space in the depth of the search. We now show how to transform one of them, Recursive Best-First Search, or RBFS (Korf, 1993), into an anytime algorithm. Besides showing how to create a linear-space anytime heuristic search algorithm, this helps to illustrate the generality of our approach by showing that another weighted heuristic search algorithm can be transformed into an anytime heuristic search algorithm in a similar way, by continuing the weighted search after the first solution is found.

We begin this section with a brief review of the RBFS algorithm. Then we consider two approaches to using a weighted evaluation function with RBFS, one that has been studied before and an alternative that we show has some advantages. Finally, we discuss how to transform Weighted RBFS into an Anytime Weighted RBFS algorithm. We use the Fifteen Puzzle as a test domain, which is a larger version of the sliding-tile puzzle that A* cannot solve optimally because of memory limitations. Because RBFS saves memory by not storing all generated nodes, it is slowed by excessive node regenerations in solving graph-search problems with many duplicate paths. As a result, RBFS is not effective (in terms of time efficiency) for either STRIPS planning or multiple sequence alignment.

## 3.1 Recursive Best-First Search (RBFS)

Recursive best-first search, or RBFS (Korf, 1993), is a general heuristic search algorithm that expands frontier nodes in best-first order, but saves memory by determining the next node to expand using stack-based backtracking instead of by selecting nodes from an Open list. The stack contains all nodes along the path from the start node to the node currently being visited, plus all siblings of each node on this path. Thus the memory complexity of RBFS is $O(db)$, where $d$ is the depth of the search and $b$ is the branching factor.

RBFS is similar to a recursive implementation of depth-first search, with the difference that it uses a special condition for backtracking that ensures that nodes are expanded (for the first time) in best-first order. Instead of continuing down the current path as far as possible, as in ordinary depth-first search, RBFS keeps track of the $f$-cost of the best alternative path available from any ancestor of the current node, which is passed as an argument to the recursive function. If the $f$-cost of the current path exceeds this threshold, called the *local cost threshold*, the recursion unwinds back to the alternative path. As the recursion unwinds, RBFS keeps track of the $f$-cost of the best unexpanded node on the frontier of the forgotten subtree by saving it in the stored value $F(n)$. These stored values, one for each node $n$ on the stack, are used by RBFS to decide which path to expand next at any point in the search. Because $F(n)$ is the least $f$-cost of any unexpanded node on the frontier of the subtree rooted at node $n$, these stored values can be propagated to successor nodes during successor generation. If a node has been previously expanded, its (propagated) stored value will be greater than its static evaluation, and RBFS uses this fact to detect previously expanded nodes and regenerate subtrees efficiently.

Among the advantages of RBFS, Korf points out that it expands nodes in best-first order even when the evaluation function is nonmonotonic. To illustrate a nonmonotonic evaluation function, he considers RBFS using a weighted evaluation function.

## 3.2 Weighted RBFS

Like A*, RBFS can use a weighted heuristic to trade off solution quality for search time. Algorithm 2 gives the pseudocode for the recursive function of RBFS using a weighted evaluation function. This is the same RBFS algorithm described by Korf, although the notation is slightly adjusted to show that the weighted values $F'$ are stored on the stack instead of the unweighted values $F$, and the local cost threshold $B'$ is a weighted value. When RBFS is initially invoked, its three arguments are the start node, the (weighted) evaluation of the start node, and a cost threshold of infinity. Using a weighted evaluation function, RBFS expands nodes (for the first time) in order of the weighted evaluation function, $f'$, instead of in order of the unweighted evaluation function, $f$. Korf (1993) considers this approach to Weighted RBFS and presents an empirical study of the tradeoff it offers between search time and solution quality.

To motivate another approach to weighted heuristic search using RBFS, we introduce a distinction between two search frontiers maintained by RBFS. The stored values, denoted $F(n)$ or $F'(n)$, keep track of the best unexpanded node on the frontier of the subtree rooted at a node $n$ on the stack. We call this a *virtual frontier* because RBFS does not actually store this frontier in memory, but uses these stored values to represent and regenerate the frontier. We introduce the term *stack frontier* to refer to the frontier that RBFS actually

---

**Algorithm 2**: RBFS (using a weighted evaluation function)

---

**Input**: A node $n, F'(n)$, and a threshold $B'$
**begin**
  **if** $n$ is a goal node **then** output solution path and exit algorithm
  **if** $Successors(n) = \emptyset$ **then return** $\infty$
  **foreach** $n_i \in Successors(n), i = 1, 2, \cdots, |Successors(n)|$ **do**
    $g(n_i) \leftarrow g(n) + c(n, n_i)$, $f'(n_i) \leftarrow g(n_i) + w \times h(n_i)$
    **if** $f'(n) < F'(n)$ **then** $F'(n_i) \leftarrow \max\{F'(n), f'(n_i)\}$
    **else** $F'(n_i) \leftarrow f'(n_i)$
  sort $n_i$ in increasing order of $F'(n_i)$
  **if** $|Successors(n)| = 1$ **then** $F'(n_2) \leftarrow \infty$
  **while** $F'(n_1) < \infty$ **and** $F'(n_1) \leq B'$ **do**
    $F'(n_1) \leftarrow$ RBFS$(n_1, F'(n_1), \min\{B', F'(n_2)\})$
    insert $n_1$ in sorted order of $F'(n_i)$
  **return** $F'(n_1)$
**end**

---

stores in memory. The stack frontier consists of the nodes on the stack that do not have successor nodes on the stack.

In weighted heuristic search, the weighted evaluation function is used to determine the order in which to expand nodes on the frontier of the search. In the approach to Weighted RBFS shown in Algorithm 2, which is the approach adopted by Korf (1993), the weighted evaluation function is used to select the order in which to expand nodes on the virtual frontier. Because this virtual frontier is the same frontier that is maintained in memory by Weighted A*, using an Open list, this approach to Weighted RBFS expands nodes in the same order as Weighted A* (disregarding tie breaking and node regenerations).

Algorithm 3 shows the pseudocode of an alternative approach to weighted heuristic search using RBFS. Like the approach shown in Algorithm 2, it uses a weighted evaluation function and continues to expand a solution path as long as the weighted evaluation of the currently-expanding node is not greater than the weighted evaluation of any sibling of one of the nodes along this path. The difference is that instead of backing up the least weighted evaluation $f'$ of any unexpanded node in the subtree rooted at node $n$ and storing it in $F'(n)$, Algorithm 3 backs up the least unweighted evaluation $f$ of any unexpanded node, and stores it in $F(n)$. If $f(n)$ is an admissible evaluation function, then $F(n)$ is a lower bound on the cost of the best solution that can be found in the subtree rooted at $n$. It follows that $H(n) = F(n) - g(n)$ is an improved admissible heuristic for node $n$. Therefore, Algorithm 3 can use the weighted evaluation $g(n) + w \times H(n) = g(n) + w \times (F(n) - g(n))$ to determine the order in which to expand nodes. In this approach to Weighted RBFS, nodes are expanded in best-first order of the weighted evaluation of nodes on the stack frontier, instead of in order of the weighted evaluation of nodes on the virtual frontier.

RBFS is a general algorithmic scheme that can use different evaluation functions. Thus, even when it uses a weighted evaluation function, Korf refers to it as RBFS. To make it easier to distinguish between these algorithms, we introduce the name WRBFS to refer to the alternative approach to weighted heuristic search based on RBFS that we propose. WRBFS expands nodes on the stack frontier in best-first order of the evaluation function

---

**Algorithm 3**: WRBFS

---

**Input**: A node $n, F(n)$, and a threshold $B'$

**begin**

  **if** $n$ is a goal node **then** output solution path and exit algorithm

  **if** $Successors(n) = \emptyset$ **then return** $\infty$

  **foreach** $n_i \in Successors(n), i = 1, 2, \cdots, |Successors(n)|$ **do**

    $g(n_i) \leftarrow g(n) + c(n, n_i),\ f(n_i) \leftarrow g(n_i) + h(n_i)$

    **if** $f(n) < F(n)$ **then** $F(n_i) \leftarrow \max\{F(n), f(n_i)\}$

    **else** $F(n_i) \leftarrow f(n_i)$

  sort $n_i$ in increasing order of $F(n_i)$

  **if** $|Successors(n)| = 1$ **then** $F(n_2) \leftarrow g(n_2) \leftarrow \infty$

  **while** $F(n_1) < \infty$ **and** $g(n_1) + w \times \big(F(n_1) - g(n_1)\big) \leq B'$ **do**

    $F(n_1) \leftarrow$ WRBFS$\big(n_1, F(n_1), \min\{B', g(n_2) + w \times \big(F(n_2) - g(n_2)\big)\}\big)$

    insert $n_1$ in sorted order of $F(n_i)$

  **return** $F(n_1)$

**end**

---

$F'(n) = g(n) + w \times H(n)$, instead of expanding nodes on the virtual frontier in order of the evaluation function $f'(n) = g(n) + w \times h(n)$. Since $F'(n) = g(n) + w \times H(n)$ improves during the search, it is not a static evaluation function, and this is another reason for using the name WRBFS. Note that when $w = 1$, as in unweighted RBFS, there is no difference in the behavior of these two algorithms; expanding nodes in best-first order of their evaluation on the stack frontier is equivalent to expanding nodes in best-first order of their evaluation on the virtual frontier. There is only a difference when one considers whether to apply a weight greater than 1 to the heuristic on the stack frontier or the virtual frontier.

Figure 4 compares the performance of these two approaches to Weighted RBFS in solving Korf's (1985) 100 random instances of the Fifteen Puzzle. Figure 4(a) shows the average length of the solutions found by each algorithm, using weights ranging from 1.0 to 5.0 in increments of 0.1. WRBFS finds better-quality solutions than RBFS using a weighted evaluation function and the same weight, and the difference increases with the weight. But since WRBFS can also take longer to find a solution, we consider the time-quality tradeoff offered by each algorithm. Figure 4(b), which is similar to Figure 10 in the article by Korf (1993), plots solution length versus time (measured by the number of recursive calls) in solving the same Fifteen Puzzle examples, using solution lengths ranging from 53 to 85 for both algorithms. The time-quality tradeoff offered by the two algorithms is similar, with a modest advantage for WRBFS. What is striking is that WRBFS offers a smooth time-quality tradeoff, whereas the tradeoff offered by RBFS using a weighted evaluation function is irregular. Sometimes, increasing the weight used by RBFS causes it to take longer to find a solution, instead of less time. A dramatic example is that increasing the weight from 1.0 to 1.1 causes RBFS with a weighted evaluation function to take three times longer to find a (potentially suboptimal) solution than unweighted RBFS takes to find an optimal solution.

Korf (1993) gives the reason for this irregular time-quality tradeoff. The node regeneration overhead of RBFS grows with the number of iterations of the algorithm, which is the number of times the local cost threshold increases, since each iteration requires regeneration of subtrees. There is one iteration for each distinct $f$-cost, or, in the case of RBFS using a weighted evaluation function, for each distinct $f'$-cost. The irregular time-quality tradeoff is caused by fluctuation in the number of distinct $f'$-costs as the weight increases, which leads to fluctuation in the number of iterations. The number of distinct $f'$-costs can be as many as the number of distinct pairs of $g$-cost and $h$-cost, but the actual number depends on the weight, since different pairs of $g$-cost and $h$-cost may sum to the same $f'$-cost, depending on the weight. Increasing the weight from 1.0 to 1.1, for example, significantly increases the number of distinct $f'$-costs, and this explains why RBFS using a weighted evaluation function takes longer to find a solution in this case. An advantage of using WRBFS is that the stored value of each node on the stack is the minimum $f$-cost on the frontier of the subtree rooted at that node, instead of the minimum $f'$-cost, and thus the number of iterations is not affected by any variation in the number of distinct $f'$-costs as the weight is increased. As a result, adjusting the weight creates a smoother time-quality tradeoff.

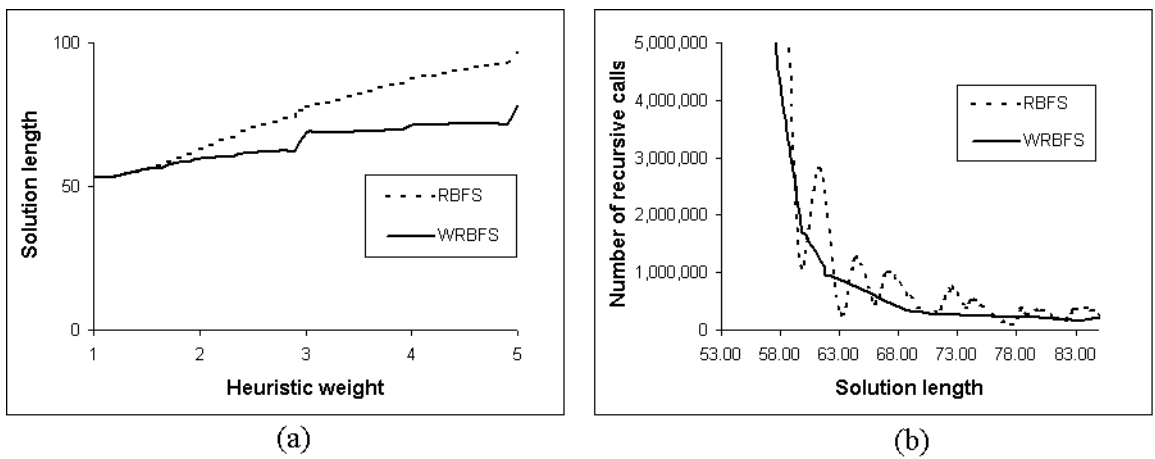

Figure 4: Comparison of RBFS using a weighted evaluation function and WRBFS in solving Korf's 100 random instances of the Fifteen Puzzle. Panel (a) shows solution quality as a function of heuristic weight. Panel (b) shows the time-quality tradeoff of each algorithm by plotting the number of recursive calls against solution quality.

Both approaches to Weighted RBFS are well-motivated and both offer a useful tradeoff between search time and solution quality. The original approach expands frontier nodes in best-first order of the weighted evaluation function $f'(n) = g(n) + w \times h(n)$, and, in this respect, it is closer to Weighted A*. But as we have seen, the alternate approach has the advantage that it allows a smoother time-quality tradeoff. As we consider how to transform each of these two approaches to Weighted RBFS into an anytime heuristic search algorithm, we will see that the alternate approach has other advantages as well.

## 3.3 Anytime Weighted RBFS

It is straightforward to transform either approach to Weighted RBFS into an anytime algorithm. Algorithm 4 shows the pseudocode of the recursive function of Anytime WRBFS. There are just a couple differences between a Weighted RBFS algorithm such as WRBFS

---

**Algorithm 4**: Anytime WRBFS

---

**Input**: A node $n$, $F(n)$, and a threshold $B'$

**begin**

  **if** $n$ is a goal node **then return** $f(n)$

  **if** $Successors(n) = \emptyset$ **then return** $\infty$

  **foreach** $n_i \in Successors(n), i = 1, 2, \cdots, |Successors(n)|$ **do**

    $g(n_i) \leftarrow g(n) + c(n, n_i)$, $f(n_i) \leftarrow g(n_i) + h(n_i)$

    **if** $n_i$ is a goal node **and** $f(n_i) < f(incumbent)$ **then**

      $incumbent \leftarrow n_i$

      save (or output) incumbent solution path

    **if** $f(n_i) \geq f(incumbent)$ **then** $F(n_i) \leftarrow \infty$

    **else if** $f(n) < F(n)$ **then** $F(n_i) \leftarrow \max\{F(n), f(n_i)\}$

    **else** $F(n_i) \leftarrow f(n_i)$

  sort $n_i$ in increasing order of $F(n_i)$

  **if** $|Successors(n)| = 1$ **then** $F(n_2) \leftarrow g(n_2) \leftarrow \infty$

  **while** $F(n_1) < f(incumbent)$ **and** $g(n_1) + w \times (F(n_1) - g(n_1)) \leq B'$ **do**

    $F(n_1) \leftarrow$ Anytime-WRBFS$\big(n_1, F(n_1), \min\{B', g(n_2) + w \times (F(n_2) - g(n_2))\}\big)$

    insert $n_1$ in sorted order of $F(n_i)$

  **return** $F(n_1)$

**end**

---

and an Anytime Weighted RBFS algorithm. Most importantly, the condition for termination is different. After the anytime algorithm finds a solution, and each time it finds an improved solution, it saves (or outputs) the solution and continues the search. As in Anytime Weighted A*, the algorithm checks whether a node is a goal node when it is generated, instead of waiting until it is expanded. It also checks whether the $f$-cost of a node is greater than or equal to an upper bound given by the $f$-cost of the incumbent solution. If so, this part of the search space is pruned. (Note that before the first solution is found, $f(incumbent)$ should be set equal to infinity, since there is not yet a finite upper bound on the optimal solution cost.) Convergence to an optimal solution is detected when the stack is empty. At this point, backtracking has determined that all branches of the tree have been searched or pruned. Proof of termination with an optimal solution follows similar logic as for Theorem 1. The suboptimality of the currently available solution is bounded by using $f(incumbent)$ as an upper bound on optimal solution cost and the least $F$-cost of any node on the stack frontier as a lower bound. (Again, the stack frontier consists of the node at the end of the current best path, plus every sibling of a node along this path.)

Figure 5(a) shows performance profiles for Anytime WRBFS, averaged over Korf's 100 random instances of the Fifteen Puzzle. Although weights of 2.0 and 1.5 offer a better time-quality tradeoff for short amounts of search time, a weight of 1.3 provides better long-term performance. Figure 5(b) shows the time (measured by the average number of recursive calls) taken by Anytime WRBFS to find optimal solutions for the Fifteen puzzle, using weights from 1.0 to 2.0 in increments of 0.1. Using weights from 1.2 to 1.4, it converges to an optimal solution more quickly than unweighted RBFS. In fact, using a weight of 1.3, Anytime WRBFS converges to an optimal solution after an average of 25% fewer recursive calls than unweighted RBFS. Although it always expands as many or more distinct nodes than unweighted RBFS, reliance on stack-based backtracking to reduce memory use means

285

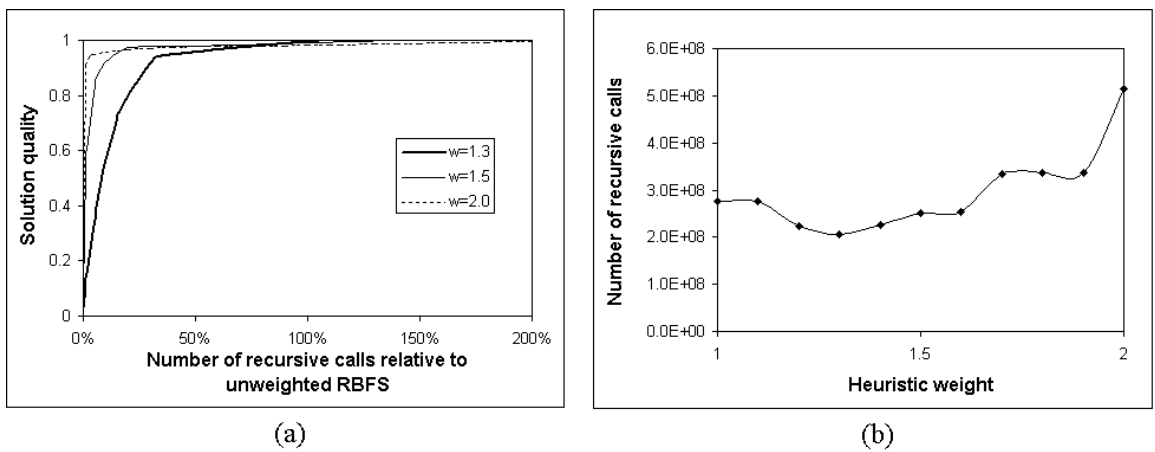

(a)                                                                                        (b)

Figure 5: Performance of Anytime WRBFS. Panel (a) shows performance profiles using
three different weights, averaged over Korf's 100 random instances of the Fifteen
Puzzle. Panel (b) shows the average number of recursive calls required to converge
to an optimal solution, using weights from 1.0 to 2.0 in increments of 0.1, and
averaged over the same 100 random instances of the Fifteen Puzzle.

that both algorithms can revisit the same nodes multiple times. Figure 5(b) shows that
the weighted heuristic used by Anytime WRBFS can reduce the number of recursive calls;
intuitively, this occurs because the greedier search strategy of weighted heuristic search
tends to delay and reduce backtracking. Of course, if the weight is increased enough, the
number of distinct node expansions increases and eventually the number of recursive calls
also increases, as Figure 5(b) shows. Nevertheless, the demonstration that a small weight
can sometimes improve efficiency in finding optimal solutions is interesting.

For comparison, Figure 6(a) shows performance profiles for a version of Anytime Weighted
RBFS that is based on RBFS using a weighted evaluation function, which is the original
approach to Weighted RBFS. In this case, the performance profile of Anytime Weighted
RBFS using a weight of 1.3 is dominated by its performance profiles using weights of 1.5
and 2.0. Figure 6(b) shows the average number of recursive calls taken by this version of
Anytime Weighted RBFS to find optimal solutions for the Fifteen puzzle, using the same
range of weights from 1.0 to 2.0. To ensure fair comparison, we implemented this version
of Anytime Weighted RBFS so that it saves the admissible $F(n)$ values in addition to the
non-admissible $F'(n)$ values, and uses the $F(n)$ values to prune branches of the search tree
and detect convergence to an optimal solution, instead of using the static evaluation $f(n)$.
Nevertheless, this version of Anytime Weighted RBFS converges much more slowly. The
scale of the $y$-axis in Figure 6(b) is an order of magnitude greater than in Figure 5(b),
and this reflects the fact that Anytime Weighted RBFS based on this version of Weighted
RBFS takes an order of magnitude longer to converge to an optimal solution than Anytime
WRBFS, using the same weight.

Fluctuations in the length of time until convergence in Figure 6(b) are caused by dif-
ferences in the number of distinct $f'$-costs as the weight increases, causing differences in

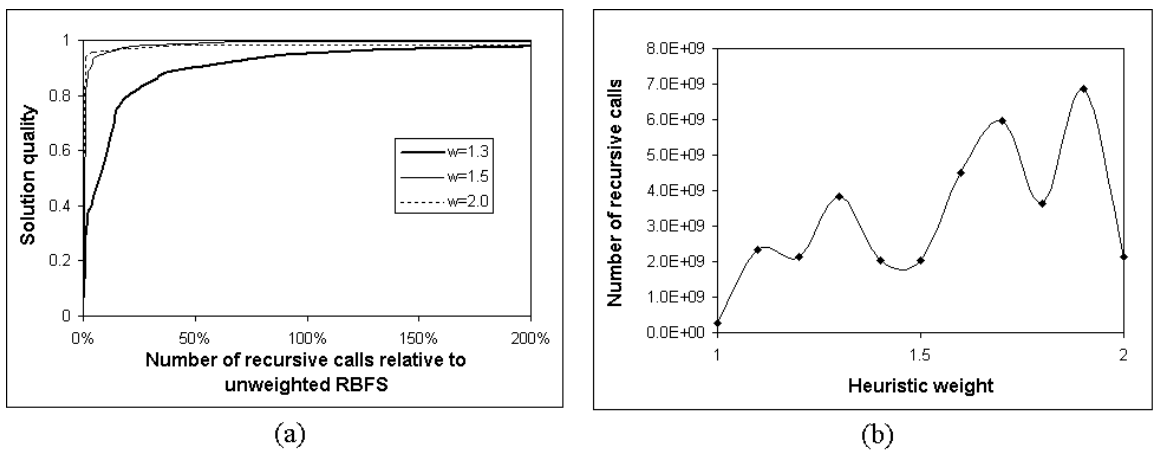

(a)                                    (b)

Figure 6: Performance of Anytime Weighted RBFS based on RBFS using a weighted evalu-
ation function. Panel (a) shows performance profiles using three different weights,
averaged over Korf's 100 random instances of the Fifteen Puzzle. Panel (b) shows
the average number of recursive calls to converge to an optimal solution, using
weights from 1.0 to 2.0 in increments of 0.1, averaged over the same instances.
The scale of the $y$-axis is an order of magnitude greater than for Figure 5(b).

the number of iterations and resulting fluctuations in node regeneration overhead. This is
similar to what we observed earlier of the performance of this approach to Weighted RBFS.
But it does not explain the very large difference in the time it takes each algorithm to con-
verge. There are at least two reasons why Anytime WRBFS converges much faster. One
is more efficient backtracking behavior. Because Anytime WRBFS expands nodes in order
of a weighted evaluation function on the stack frontier, instead of in order of a weighted
evaluation function on the virtual frontier, it searches more greedily at deeper levels on the
stack before backtracking to shallower levels. Since it is more computationally expensive
to regenerate the large subtrees that are rooted at shallower nodes on the stack than the
smaller subtrees that are rooted at deeper nodes, this bias towards backtracking at deeper
levels before backtracking to shallower levels contributes to improved convergence time.

Another reason that Anytime WRBFS converges much faster is that it is more effective
in improving the lower bound on optimal solution cost. As we pointed out earlier, anytime
heuristic search often finds what turns out to be an optimal solution relatively quickly,
and spends most of its time proving that the solution is optimal, which corresponds to
improving a lower bound. In both versions of Anytime Weighted RBFS, the lower bound is
the minimum of the $F(n)$ values stored on the stack frontier. Although an anytime search
algorithm based on the original version of Weighted RBFS is guaranteed to improve the
$F'(n)$ value of a subtree rooted at node $n$ each iteration, it may or may not improve the $F(n)$
value (which we assume it also stores). By contrast, an anytime search algorithm based
on WRBFS is guaranteed to improve the $F(n)$ value each iteration. Because it improves
admissible $F(n)$ values, instead of weighted $F'(n)$ values, Anytime WRBFS is more effective
in improving the lower bound on optimal solution cost, leading to faster convergence.

## 4. Related Work

In this section, we consider some closely-related work. We begin by considering a variant of Anytime A* that has been recently proposed. Then we discuss the relationship between Anytime A* and other variants of A* that, directly or indirectly, also allow a tradeoff between search time and solution quality.

### 4.1 Anytime Repairing A*

Likhachev, Gordon, and Thrun (2004) have recently introduced an interesting variant of Anytime A*, called *Anytime Repairing A** (or ARA*), and have shown that it is very effective for real-time robot path planning. Their approach follows our approach to creating an Anytime A* algorithm in many respects. It uses Weighted A* to find an approximate solution relatively quickly, and continues the weighted search to find a sequence of improved solutions until convergence to an optimal solution. However, it introduces two extensions to improve performance. First, after each solution is found, it decreases the weight before continuing the search. Second, it uses a technique to limit node reexpansions. The first of these extensions, decreasing the weight as new solutions are found, is easy to consider independently of the other, and can also be easily combined with Anytime Weighted A* (AWA*), and so we consider it first.

**Decreasing the weight** In our experiments, we used a weighted heuristic with a weight that did not change during the search. We chose this approach because of its simplicity. Likhachev et al. (2004) argue that better performance is possible by gradually decreasing the weight during search. After each new solution is found, ARA* decreases the weight by a small, fixed amount, and continues the search. Experimental results show that this approach leads to improved performance in their robot path-planning domains.

Of course, relative performance can depend on the initial weight and not simply on whether the weight remains fixed or decreases. Likhachev et al. report three experimental comparisons of AWA* and ARA*. In one, they set the initial weight to 3; in another, they set the initial weight to 10; in the third, they set the initial weight to 30. These weights are higher than those we found worked well for our test problems. In their experiments, AWA* never changes this initial weight whereas ARA* decreases it as new solutions are found. If the initial weight is set too high, this might explain why decreasing it improves performance. It could also be that setting the initial weight high and gradually decreasing it is the most effective approach for the robot path-planning problems they consider, and for similar problems. Even so, it does not follow that it is the best approach for all problems.

We compared Anytime Weighted A* and Anytime Repairing A* in solving the STRIPS planning problems used as a testbed in Section 2.4.2. Although Likhachev et al. did not use upper bounds to reduce the size of the Open list in their implementation of ARA*, it is easy to do so and we included this enhancement in our implementation of ARA* in order to ensure fair comparison. In addition to decreasing the weight, ARA* uses a special "repairing" technique to limit node reexpansions. Since this is an independent idea, we implemented a version of AWA* that decreases the weight during search but does not use this special technique for limiting node reexpansions. By itself, decreasing the weight during search only requires recalculating $f'$-costs and reordering the Open list whenever the weight is changed. Table 3 compares the performance of AWA* and ARA* when both use an initial weight of 2.0 and decrease the weight by 0.1 after each new solution is found. For all planning instances except Logistics-6, Freecell-3 and Psr-46, there is no significant difference in their performance or any significant difference between their performance and the performance of AWA* with a fixed weight of 2.0 in solving the same instances. (See Table 1.) For these STRIPS planning problems and using this initial weight, gradually decreasing the weight does not improve performance. Of course, it could improve performance for other problems. In that case, we note that it is easy to decrease the weight used by AWA* without implementing the full ARA* algorithm.

| Instances | AWA* (weight = 2, step = 0.1) | | | | ARA* (weight = 2, step = 0.1) | | | |
|---|---|---|---|---|---|---|---|---|
| | Stored | Exp | Opt % | Secs | Stored | Exp | Opt % | Secs |
| Blocks-8 | 41,166 | 40,727 | 0.2% | 3.9 | 41,166 | 42,141 | 0.2% | 3.9 |
| Logistics-6 | 254,412 | 254,390 | 6.2% | 3.6 | 312,438 | 364,840 | 87.1% | 5.0 |
| Satellite-6 | 2,423,547 | 2,423,489 | 14.3% | 138.6 | 2,423,547 | 2,428,325 | 14.2% | 138.4 |
| Freecell-3 | 2,695,321 | 2,698,596 | 1.7% | 155.3 | 4,115,032 | 5,911,849 | 68.7% | 317.2 |
| Psr-46 | 7,148,048 | 7,171,557 | 69.0% | 345.7 | 7,143,912 | 11,888,700 | 35.3% | 567.7 |
| Depots-7 | 7,773,830 | 7,762,783 | 0.5% | 247.4 | 7,771,780 | 7,823,005 | 0.5% | 247.1 |
| Driverlog-11 | 6,763,379 | 6,693,441 | 4.7% | 283.3 | 6,698,404 | 6,771,651 | 1.3% | 281.3 |
| Elevator-12 | 12,734,636 | 12,825,980 | 98.7% | 561.3 | 12,736,328 | 12,843,441 | 97.8% | 559.7 |

Table 3: Comparison of AWA* (with decreasing weight) and ARA* on eight benchmark problems from the biennial Planning Competitions.

Another potential advantage of decreasing the weight, as Likhachev et al. point out, is that it provides a different way of bounding the suboptimality of a solution. For any solution found by Weighted A* using a weight of $w$, one has the error bound, $\frac{f(incumbent)}{f^*} \leq w$. Note that decreasing this bound requires decreasing the weight during the search.

In Section 2.3.2, we defined a different error bound, $\frac{f(incumbent)}{f^*} \leq \frac{f(incumbent)}{f^L}$, where $f^L$ denotes the least $f$-cost of any currently open node on the frontier. An advantage of this error bound is that it decreases even if the weight remains fixed during the search. An additional advantage is that it is a tighter bound. Let $n^L$ denote an open node with $f(n^L) = f^L$. Because $h(incumbent) = 0$ and $incumbent$ was expanded before $n^L$, we know that

$$f(incumbent) = f'(incumbent) \leq g(n^L) + w \times h(n^L).$$

Therefore,

$$\frac{f(incumbent)}{f^L} \leq \frac{g(n^L) + w \times h(n^L)}{g(n^L) + h(n^L)} < \frac{w \times (g(n^L) + h(n^L))}{g(n^L) + h(n^L)} = w,$$

where the strict inequality follows from the assumptions that $w > 1$ and $g(n^L) > 0$.

Although ARA* performs about the same as AWA* in solving five of the eight planning problems, it performs worse in solving the other three: Logistics-6, Freecell-3, and Psr-46. Comparing ARA* to AWA* when both have the same initial weight and decrease the weight in the same way shows that this deterioration in performance is not caused by decreasing the weight. We consider next the technique used by ARA* for limiting node reexpansions.

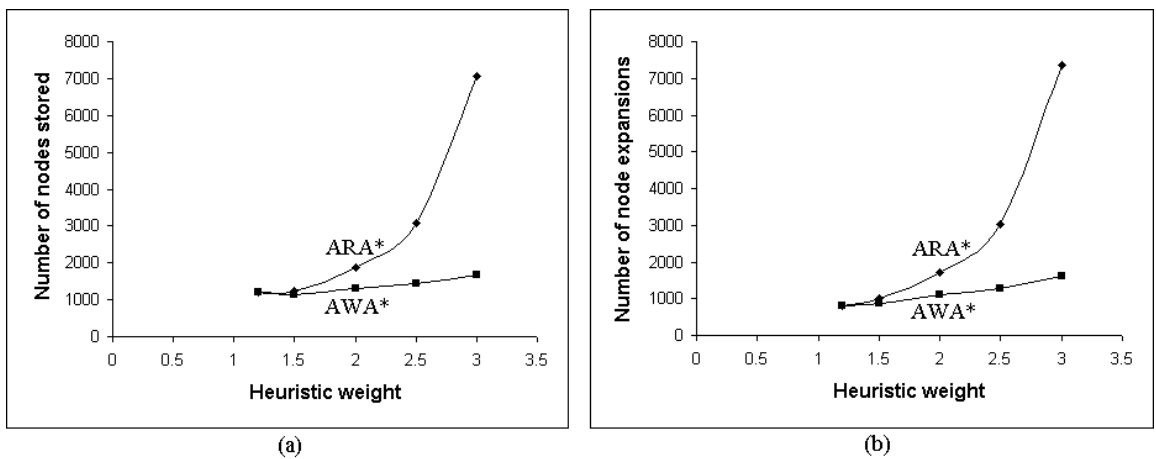

Figure 7: Comparison of AWA* and ARA* in solving all instances of the Eight Puzzle. Panel (a) shows the average number of nodes stored and panel (b) shows the average number of node expansions, both as a function of the initial weight.

**Limiting node reexpansions** As discussed before, a complication that Anytime Weighted A* inherits from Weighted A* is that a weighted heuristic is typically inconsistent. This means it is possible for a node to have a higher-than-optimal $g$-cost when it is expanded. If a better path to the same node is later found, the node is reinserted in the Open list so that the improved $g$-cost can be propagated to its descendants when the node is reexpanded. As a result, both Weighted A* and Anytime WA* can expand the same node multiple times.

Likhachev et al. note that the error bound for Weighted A* remains valid even if node reexpansions are not allowed. Since their ARA* algorithm performs a series of Weighted A* searches with decreasing weights, they reason that if ARA* postpones node reexpansions until the current iteration of Weighted A* finishes and the weight is decreased, this will create a more efficient Anytime A* algorithm. When ARA* finds a better path to an already-expanded node, it inserts the node into a list called INCONS in order to delay node reexpansion. When a solution is found and the weight is decreased, ARA* moves all nodes in the INCONS list to the Open list and resumes the search.

This technique for limiting node reexpansions may improve search performance for robot path planning and similar search problems, especially when using large weights. But the relative performance of AWA* and ARA* in solving the Logistics-6, Freecell-3, and Psr-46 planning instances raises a question about whether it always improves performance. For further comparison, Table 7 shows the average performance of AWA* and ARA* in solving all instances of the Eight Puzzle. Each algorithm has the same initial weight. AWA* never changes the initial weight while ARA* reduces it in increments of 0.1 as new solutions are found. A larger weight makes the heuristic more inconsistent and increases the likelihood that AWA* will reexpand nodes. Yet the results show that the larger the initial weight, the more nodes ARA* expands relative to AWA*, and the difference is dramatic. When the initial weight is set to 3.0, ARA* expands more than four times more nodes than AWA*.

In many cases, it takes longer for ARA* to find an initial suboptimal solution than it takes unweighted A* to find an optimal solution.

One reason for this result is that limiting node reexpansions can cause ARA* to expand more distinct nodes. (The fact that ARA* stores as well as expands more nodes, as shown in Figure 7, indicates that it expands and generates more distinct nodes.) Limiting node reexpansions can lead to expansion of more distinct nodes because it blocks improvement of all paths that pass through any node stored in the INCONS list. By blocking improvement of these paths, it can prevent better solutions from being found. One possibility is that the solution found by Weighted A* passes through a node that is stored in the INCONS list, which means that reexpansion and propagation of its improved $g$-cost is postponed. In that case, the $f'$-cost of the solution is greater than it would be if reexpansion of the node was allowed. Another possibility is that a potentially better solution than the one found by Weighted A* passes through a node in the INCONS list, and therefore is not discovered because its improvement is blocked. Either way, the solution found by Weighted A* when it does not allow node reexpansions can have a greater $f'$-cost than if node reexpansions are allowed. Because Weighted A* must expand all nodes with an $f'$-cost less than the $f'$-cost of the solution it finds, more distinct nodes can be expanded whenever limiting node reexpansions prevents Weighted A* from finding a better solution. As Figure 7 shows, this effect becomes more pronounced as increasing the weight increases the likelihood that the first time ARA* expands a node, its $g$-cost is suboptimal.

Our results show that this effect does not occur for all search problems, at least to the same degree. It seems to occur primarily for search problems with relatively sparse solutions, such as the sliding-tile puzzle and the Logistics and Freecell planning domains. When solutions are sparse, it is easier for *all* nodes that lead to a good solution to be expanded with a higher-than-optimal $g$-cost, and thus more likely for Weighted A* to find a solution that is worse than it would have found if it allowed node reexpansions. For search problems with a huge number of solutions of equal or almost equal cost, limiting node reexpansions in this way is less likely to cause the same problem. The robot path-planning problems considered by Likhachev *et al.* are examples of this kind of search problem, and thus the impressive results they report are not inconsistent with our observations.

There is yet another way in which limiting node reexpansions sometimes makes search performance worse. So far, we have considered search problems where ARA* expands more distinct nodes than AWA*. But for the Psr-46 planning instance, ARA* expands many more nodes than AWA*, but does not store more nodes. This indicates that ARA* does not expand more distinct nodes than AWA*. Instead, it performs more node reexpansions. How is this possible when ARA* explicitly limits node reexpansions? It turns out that limiting node reexpansions in the way that ARA* does can sometimes lead to more node reexpansions. By the time ARA* decreases its weight and reexpands a node to propagate improved path information, the reexpanded node can have many more descendants in the explicit search graph than it did when an improved path to the node was originally found. As a result, many more nodes may need to be reexpanded to propagate the improved path information. Again, this does not always happen. But the behavior of ARA* in solving Psr-46 illustrates this possibility.

Figure 7(b) compares the average number of nodes expanded by ARA* and AWA* in solving all instances of the Eight Puzzle, but it does not show CPU time. With an initial weight of 3, ARA* expands about 4.5 times more nodes than AWA*. The difference in CPU time is actually greater. ARA* takes 7 times longer to solve these problems than AWA*, on average. One reason for this is the extra time overhead for recalculating $f'$-costs and reordering the Open list every time the weight is decreased. This time overhead is negligible for the STRIPS planning problems compared to the much greater overhead for domain-independent node generation and heuristic calculation. But for the sliding-tile puzzle, node generation and heuristic calculation are so fast that the time overhead for recalculating $f'$-costs and reordering the Open list has a noticeable effect in slowing the search. This is another example of how the relative performance of ARA* and AWA* can vary with the search problem.

In summary, the idea of decreasing the weight during search can be used independently of the technique for limiting node reexpansions. Although gradually decreasing the weight did not improve performance for our test problems, it could improve performance for other problems. However, the additional overhead for recalculating $f'$-costs and reordering the Open list should be considered. The technique for limiting node reexpansions can also help, but should be used with caution. For some problems, we have shown that it can actually cause significantly more node reexpansions or expansion of more distinct nodes. For other problems, it does not have a negative effect. Although it did not show a clear benefit in our test domains, it could improve performance for robot path planning and similar problems with many close-to-optimal solutions, especially when using a large weight.

## 4.2 Real-time A*

An anytime approach to heuristic search is effective for real-time search problems where not enough time is available to search for an optimal solution. Previous work on time-limited heuristic search adopts the model of Korf's Real-time A* algorithm (RTA*) which assumes that search is interleaved with execution (Korf, 1990). After searching for a bounded amount of time, the best next action is chosen and the search-act cycle repeats until the goal state is reached. Similar examples of this real-time search strategy include DTA* (Russell & Wefald, 1991), BPS (Hansson & Mayer, 1990) and $k$-best (Pemberton, 1995). Because real-time search algorithms commit to actions before finding a complete solution, they cannot find optimal solutions. In contrast, we assume that a search phase precedes an execution phase and that the output of the search is a complete solution. In other words, real-time search algorithms try to find the *best next decision* under a time constraint, whereas our anytime approach tries to find the *best complete solution* under a time constraint.

## 4.3 Depth-first branch and bound and Iterative-Deepening A*

Depth-first branch-and-bound (DFBnB) search algorithms are very effective for tree-search problems, especially those that have many solutions at the same depth, such as the traveling salesman problem. For such problems, DFBnB has the behavior of an anytime algorithm. It quickly finds a solution that is suboptimal, and then continues to search for improved solutions until an optimal solution is found. It even uses the cost of the best solution found so far as an upper bound to prune the search space.

A search technique that combines elements of DFBnB and A* is Iterative-deepening A* or IDA* (Korf, 1985). It is well-known that IDA* performs poorly on problems with real-valued edge costs such as the traveling salesman where almost all nodes have distinct $f$-costs. For such problems, it may expand only one new node each iteration. To prevent excessive iterations and node regenerations, several variants of IDA* have been developed that set successive thresholds higher than the minimum $f$-cost that exceeded the previous threshold (Sarkar, Chakrabarti, Ghose, & DeSarkar, 1991; Rao, Kumar, & Korf, 1991; Wah & Shang, 1994). As a result, the first solution found is not guaranteed to be optimal, although it has bounded error. After finding an initial solution, these algorithms revert to DFBnB search to ensure eventual convergence to an optimal solution. This approach to reducing node regenerations in IDA* has the side-effect of creating an anytime algorithm, although one that is only effective for problems for which IDA* and DFBnB are effective, which are typically tree-search problems.

## 4.4 Bidirectional A*

Another search technique that has the side-effect of creating an anytime algorithm is Bidirectional A* (Kaindl & Kainz, 1997). In this approach, two simultaneous A* searches are performed, one from the start state to the goal, and the other from the goal to the start state. When the two search frontiers meet at a node, the two partial solutions are combined to create a complete solution. Typically, the first solution found is suboptimal and the search must be continued to find an optimal solution. Thus, a bidirectional search strategy has the side-effect of finding a succession of improved solutions before convergence to an optimal solution. In fact, the convergence test used by Bidirectional A* to detect an optimal solution is similar to the convergence test used by Anytime WA*: an incumbent solution of cost $f(incumbent)$ is optimal if there is no unexpanded node with an $f$-cost less than $f(incumbent)$ in one of the two directions of the search, that is, if one of the two Open lists is empty. An interesting possibility for improving bidirectional search is to use Anytime A* (instead of standard A*) to search in both directions.

## 4.5 Local-search variants of A*

An important class of anytime search algorithms relies on local search in some form to iteratively improve a solution. Although local-search algorithms cannot guarantee convergence to an optimal solution, they scale much better than systematic search algorithms. There have been a couple attempts to improve the scalability of A* by integrating it with some form of local search. Ratner and Pohl (1986) propose two local-search variants of A* that improve a suboptimal solution path by making local searches around segments of the path. For example, *Joint A*** divides an initial suboptimal solution path into segments, and, for each segment, uses A* to search for a better path between the start and end states of the segment, to reduce overall solution cost. Ikeda et al. (1999) propose a *K-group* A* algorithm for multiple sequence alignment that performs A* on groups of sequences, instead of individual sequences, in order to reduce search complexity when the number of sequences is too large to find optimal alignments. By varying the groupings, a local-search algorithm is created that gradually improves an alignment in a computationally feasible way (Zhou & Hansen, 2004). But, like other local-search algorithms, these local-search variants of A* do not guarantee convergence to an optimal solution.

## 5. Conclusion

We have presented a simple approach for converting a heuristic search algorithm such as A* into an anytime algorithm that offers a tradeoff between search time and solution quality. The approach uses weighted heuristic search to find an initial, possibly suboptimal solution, and then continues to search for improved solutions until convergence to a provably optimal solution. It also bounds the suboptimality of the currently available solution.

The simplicity of the approach makes it very easy to use. It is also widely applicable. Not only can it be used with other search algorithms that explore nodes in best-first order, such as RBFS, we have shown that it is effective in solving a wide range of search problems. As a rule, it is effective whenever a suboptimal solution can be found relatively quickly using a weighted heuristic, and finding a provably optimal solution takes much longer. That is, it is effective whenever weighted heuristic search is effective. If the weight is chosen appropriately, we have shown that this approach can create a search algorithm with attractive anytime properties without significantly delaying convergence to a provably optimal solution. We conclude that anytime heuristic search provides an attractive approach to challenging search problems, especially when the time available to find a solution is limited or uncertain.

### Acknowledgments

We are grateful to Shlomo Zilberstein for encouragement of this work, especially in its early stages. We appreciate the very helpful comments and suggestions of the anonymous reviewers, which led to several improvements of the paper. We also thank Rich Korf for helpful feedback about the RBFS algorithm. This research was supported in part by NSF grant IIS-9984952 and NASA grant NAG-2-1463.

## References

Bagchi, A., & Mahanti, A. (1983). Search algorithms under different kinds of heuristics – a comparative study. *Journal of the ACM, 30*(1), 1–21.

Bagchi, A., & Srimani, P. (1985). Weighted heuristic search in networks. *Journal of Algorithms, 6*, 550–576.

Bonet, B., & Geffner, H. (2001). Planning as heuristic search. *Artificial Intelligence, 129*(1), 5–33.

Carillo, H., & Lipman, D. (1988). The multiple sequence alignment problem in biology. *SIAM Journal of Applied Mathematics, 48*(5), 1073–1082.

Chakrabarti, P., Ghosh, S., & DeSarkar, S. (1988). Admissibility of AO* when heuristics overestimate. *Artificial Intelligence, 34*(1), 97–113.

Davis, H., Bramanti-Gregor, A., & Wang, J. (1988). The advantages of using depth and breadth components in heuristic search. In Ras, Z., & Saitta, L. (Eds.), *Methodologies for Intelligent Systems 3*, pp. 19–28.

Dechter, R., & Pearl, J. (1985). Generalized best-first search strategies and the optimality of A*. *Journal of the ACM*, *32*(3), 505–536.

Gasching, J. (1979). *Performance measurement and analysis of certain search algorithms*. Ph.D. thesis, Carnegie-Mellon University. Department of Computer Science.

Hansen, E., Zilberstein, S., & Danilchenko, V. (1997). Anytime heuristic search: First results. Tech. rep. 97-50, Univ. of Massachusetts/Amherst, Dept. of Computer Science.

Hansen, E., & Zilberstein, S. (2001). LAO*: A heuristic search algorithm that finds solutions with loops. *Artificial Intelligence*, *129*(1–2), 35–62.

Hansson, O., & Mayer, A. (1990). Probabilistic heuristic estimates. *Annals of Mathematics and Artificial Intelligence*, *2*, 209–220.

Harris, L. (1974). The heuristic search under conditions of error. *Artificial Intelligence*, *5*(3), 217–234.

Hart, P., Nilsson, N., & Raphael, B. (1968). A formal basis for the heuristic determination of minimum cost paths. *IEEE Transactions on Systems Science and Cybernetics (SSC)*, *4*(2), 100–107.

Haslum, P., & Geffner, H. (2000). Admissible heuristics for optimal planning. In *Proceedings of the 5th International Conference on Artifial Intelligence Planning and Scheduling (AIPS-00)*, pp. 140–149. AAAI Press.

Ikeda, T., & Imai, T. (1994). Fast A* algorithms for multiple sequence alignment. In *Genome Informatics Workshop 94*, pp. 90–99.

Ikeda, T., & Imai, H. (1999). Enhanced A* algorithms for multiple alignments: Optimal alignments for several sequences and k-opt approximate alignments for large cases. *Theoretical Computer Science*, *210*(2), 341–374.

Kaindl, H., & Kainz, G. (1997). Bidirectional heuristic search reconsidered. *Journal of Artificial Intelligence Research*, *7*, 283–317.

Kobayashi, H., & Imai, H. (1998). Improvement of the A* algorithm for multiple sequence alignment. In *Proceedings of the 9th Workshop on Genome Informatics (GIW'98)*, pp. 120–130. Universal Academy Press, Inc.

Koll, A., & Kaindl, H. (1992). A new approach to dynamic weighting. In *Proceedings of the 10th European Conference on Artificial Intelligence (ECAI-92)*, pp. 16–17. John Wiley and Sons.

Korf, R. (1985). Depth-first iterative deepening: An optimal admissible tree search. *Artificial Intelligence*, *27*(1), 97–109.

Korf, R. (1990). Real-time heuristic search. *Artificial Intelligence*, *42*(2–3), 197–221.

Korf, R. (1993). Linear-space best-first search. *Artificial Intelligence*, *62*(1), 41–78.

Likhachev, M., Gordon, G., & Thrun, S. (2004). ARA*: Anytime A* with provable bounds on sub-optimality. In *Advances in Neural Information Processing Systems 16: Proceedings of the 2003 Conference (NIPS-03)*. MIT Press.

Long, D., & Fox, M. (2003). The 3rd international planning competition: Results and analysis. *Journal of Artificial Intelligence Research*, *20*, 1–59.

Pearl, J. (1984). *Heuristics: Intelligent search strategies for computer problem solving*. Addison-Wesley.

Pearl, J., & Kim, J. (1982). Studies in semi-admissible heuristics. *IEEE Transactions on Pattern Analysis and Machine Intelligence*, *PAMI-4*(4), 392–399.

Pemberton, J. (1995). *k*-best: A new method for real-time decision making. In *Proceedings of the 14th International Joint Conference on Artificial Intelligence (IJCAI-95)*, pp. 227–233. Morgan Kaufmann.

Pohl, I. (1970a). First results on the effect of error in heuristic search. *Machine Intelligence*, *5*, 219–236.

Pohl, I. (1970b). Heuristic search viewed as path finding in a graph. *Artificial Intelligence*, *1*(3), 193–204.

Pohl, I. (1973). The avoidance of (relative) catastrophe, heuristic competence, genuine dynamic weighting and computational issues in heuristic problem-solving. In *Proceedings of the 3rd International Joint Conference on Artificial Intelligence (IJCAI-73)*, pp. 12–17. Morgan Kaufmann.

Rao, V., Kumar, V., & Korf, R. (1991). Depth-first vs. best-first search. In *Proceedings of the 9th National Conference on Artificial Intelligence (AAAI-91)*, pp. 434–440. AAAI/MIT Press.

Ratner, D., & Pohl, I. (1986). Joint and LPA*: Combination of approximation and search. In *Proceedings of the 5th National Conference on Artificial Intelligence (AAAI-86)*, pp. 173–177. AAAI/MIT Press.

Russell, S., & Wefald, E. (1991). *Do the Right Thing: Studies in Limited Rationality*. MIT Press.

Sarkar, U., Chakrabarti, P., Ghose, S., & DeSarkar, S. (1991). Reducing reexpansions in iterative-deepening search by controlling cutoff bounds. *Artificial Intelligence*, *50*(2), 207–221.

Shimbo, M., & Ishida, T. (2003). Controlling the learning process of real-time heuristic search. *Artificial Intelligence*, *146*(1), 1–41.

Wah, B., & Shang, Y. (1994). A comparative study of IDA*-style searches. In *Proceedings of the 6th International Conference on Tools with Artificial Intelligence (ICTAI-94)*, pp. 290–296. IEEE Computer Society.

Yoshizumi, T., Miura, T., & Ishida, T. (2000). A* with partial expansion for large branching factor problems. In *Proceedings of the 17th National Conference on Artificial Intelligence (AAAI-2000)*, pp. 923–929. AAAI/MIT Press.

Zhou, R., & Hansen, E. (2002). Multiple sequence alignment using Anytime A*. In *Proceedings of the 18th National Conference on Artificial Intelligence (AAAI-02)*, pp. 975–6. AAAI/MIT Press.

Zhou, R., & Hansen, E. (2004). K-group A* for multiple sequence alignment with quasi-natural gap costs. In *Proceedings of the 16th IEEE International Conference on Tools with Artificial Intelligence (ICTAI-04)*, pp. 688–695. IEEE Computer Society.

Zilberstein, S. (1996). Using anytime algorithms in intelligent systems. *AI Magazine, 17*(3), 73–83.